\documentclass[sigconf]{acmart}

\AtBeginDocument{%
  \providecommand\BibTeX{{%
    \normalfont B\kern-0.5em{\scshape i\kern-0.25em b}\kern-0.8em\TeX}}}

\setcopyright{acmcopyright}
\copyrightyear{2019}
\acmYear{2019}
\acmDOI{10.1145/1122445.1122456}

\acmConference[London '19]{London '19: ACM Workshop on Artificial Intelligence and Security}{November 15, 2019}{London, UK}
\acmBooktitle{London '19: ACM Workshop on Artificial Intelligence and Security,
  November 15, 2019, London, UK}

\usepackage{amssymb, amsmath, amsthm, amsfonts}
\usepackage{xfrac}
\newtheorem{definition}{Definition}
\usepackage{algorithm}
\usepackage{algorithmic}

\DeclareMathOperator*{\argmax}{arg\,max}

\usepackage{microtype}
\usepackage{graphicx}
\usepackage{subfigure}
\usepackage{booktabs} 

\usepackage{amssymb, amsmath, amsthm, amsfonts}
\usepackage{xfrac}

\newlength\myindent
\newcommand\bindent{%
 \begingroup
 \setlength{\itemindent}{\myindent}
}
\newcommand\eindent{\endgroup}

\usepackage{mdwlist}

\makecompactlist{itemize}{stditemize}

\makecompactlist{enumerate}{stdenumerate}

\begin{document}

\title{A Hybrid Approach to Privacy-Preserving Federated Learning}

\author{Stacey Truex}
\email{staceytruex@gatech.edu}
\affiliation{%
  \institution{Georgia Institute of Technology}
  \streetaddress{North Ave NW}
  \city{Atlanta}
  \state{Georgia}
  \postcode{30332}
}

\author{Nathalie Baracaldo}
\email{baracald@us.ibm.com}
\affiliation{%
  \institution{IBM Research Almaden}
  \city{San Jose}
  \country{California}}

\author{Ali Anwar}
\email{Ali.Anwar2@ibm.com}
\affiliation{%
  \institution{IBM Research Almaden}
  \city{San Jose}
  \country{California}}

\author{Thomas Steinke}
\email{Thomas.Steinke@ibm.com}
\affiliation{%
  \institution{IBM Research Almaden}
  \city{San Jose}
  \country{California}}

\author{Heiko Ludwig}
\email{hludwig@us.ibm.com}
\affiliation{%
  \institution{IBM Research Almaden}
  \city{San Jose}
  \country{California}}

\author{Rui Zhang}
\email{ruiz@us.ibm.com}
\affiliation{%
  \institution{IBM Research Almaden}
  \city{San Jose}
  \country{California}}

\author{Yi Zhou}
\email{yi.zhou@ibm.com}
\affiliation{%
  \institution{IBM Research Almaden}
  \city{San Jose}
  \country{California}}


\begin{abstract}
Federated learning facilitates the collaborative training of models without the sharing of raw data. However, recent attacks demonstrate that simply maintaining data locality during training processes does not provide sufficient privacy guarantees. Rather, we need a federated learning system capable of preventing inference over both the messages exchanged during training and the final trained model while ensuring the resulting model also has acceptable predictive accuracy.
Existing federated learning approaches either use secure multiparty computation (SMC) which is vulnerable to inference or differential privacy which 
can lead to low accuracy given a large number of parties with relatively small amounts of data each. In this paper, we present an alternative approach that utilizes both differential privacy and SMC to balance these trade-offs. Combining differential privacy with secure multiparty computation enables us to reduce the growth of noise injection as the number of parties increases without sacrificing privacy while maintaining a pre-defined rate of trust. Our system is therefore a
scalable approach that protects against inference threats and produces models with high accuracy. Additionally, our system can be used to train a variety of machine learning models, which we validate with experimental results on 3 different machine learning algorithms. Our experiments demonstrate that our approach out-performs state of the art solutions.
\end{abstract}


\maketitle
\setlength{\textfloatsep}{0.5\baselineskip}

\section{Introduction}

In traditional machine learning (ML) environments, training data is centrally held by one organization executing the learning algorithm.  Distributed learning systems extend this approach by using a set of learning nodes accessing shared data or having the data sent to the participating nodes from a central node, all of which are fully trusted.  For example, MLlib from Apache Spark assumes a trusted central node to coordinate distributed learning processes~\cite{meng2016mllib}. Another  approach is the parameter server~\cite{li2014scaling}, which again requires a fully trusted central node to collect and aggregate parameters from the many nodes learning on their different datasets.

However, some learning scenarios must address less open trust boundaries, particularly when multiple organizations are involved. While a larger dataset improves the performance of a trained model, organizations often cannot share data due to legal restrictions or competition between participants.
For example, consider three hospitals with different owners serving the same city. Rather than each hospital creating their own predictive model forecasting cancer risks for their patients, the hospitals want to create a model learned over the whole patient population. However, privacy laws prohibit them from sharing their patients' data. Similarly, a service provider may collect usage data both in Europe and the United States. Due to legislative restrictions, the service provider's data cannot be stored in one central location. When creating a predictive model forecasting service usage, however, all datasets should be used.

The area of federated learning (FL) addresses these more restrictive environments wherein data holders collaborate throughout the learning process rather than relying on a trusted third party to hold data~\cite{shokri2015privacy,bonawitz2017practical}. Data holders in FL run a machine learning algorithm locally and only exchange model parameters, which are aggregated and redistributed by one or more central entities. However, this approach is not sufficient to provide reasonable data privacy guarantees. We must also consider that information can be inferred from the learning process~\cite{nasr2018comprehensive} and that information that can be traced back to its source in the resulting trained model~\cite{shokri2017membership}.

Some previous work has proposed a trusted aggregator as a way to control privacy exposure~\cite{abadi2016deep}, \cite{papernot:pate:2018}. FL schemes using Local Differential Privacy also address the privacy problem~\cite{shokri2015privacy} but entails adding too much noise to model parameter data from each node, often yielding poor performance of the resulting model.

We propose a novel federated learning system which provides formal privacy guarantees, accounts for various trust scenarios, and produces models with increased accuracy when compared with existing privacy-preserving approaches.
Data never leaves the participants and privacy is guaranteed using secure multiparty computation (SMC) and differential privacy. 
We account for potential inference from individual participants as well as the risk of collusion amongst the participating parties through a customizable trust threshold. Our \textbf{contributions} are the following:
\begin{itemize}
    \item We propose and implement an FL system providing formal privacy guarantees and models with improved accuracy compared to existing approaches.
    \item We include a tunable trust parameter which accounts for various trust scenarios while maintaining the improved accuracy and formal privacy guarantees.
    \item We demonstrate that it is possible to use the proposed approach to train a variety of ML models through the experimental evaluation of our system with three significantly different ML models: decision trees, convolutional neural networks and linear support vector machines.
    \item We include the first federated approach for the private \textit{and} accurate training of a neural network model.
\end{itemize}

The rest of this paper is organized as follows. We outline the building blocks in our system. We then discuss the various privacy considerations in FL systems followed by outlining our threat model and general system. We then provide experimental evaluation and discussion of the system implementation process. Finally, we give an overview of related work and some concluding remarks.

\section{Preliminaries}

In this section we introduce building blocks of our approach and explain how various approaches fail to protect data privacy in FL.

\subsection{Differential Privacy}

Differential privacy (DP) is a rigorous mathematical framework wherein an algorithm may be described as differentially private if and only if the inclusion of a single instance in the training dataset causes only statistically insignificant changes to the algorithm's output. For example, consider private medical information from a particular hospital. The authors in~\cite{shokri2017membership} have shown that with access to only a trained ML model, attackers can infer whether or not an individual was a patient at the hospital, violating their right to privacy. DP puts a theoretical limit on the influence of a single individual, thus limiting an attacker's ability to infer such membership. The formal definition for DP is~\cite{dwork2008differential}:
\begin{definition}[Differential Privacy]
A randomized mechanism $\mathcal{K}$ provides $(\epsilon, \delta)$-
differential privacy if for any two neighboring database $D_1$ and $D_2$
that differ in only a single entry, $\forall S \subseteq
Range(\mathcal{K})$,
\begin{equation}
\Pr(\mathcal{K}(D_1) \in S) \le e^\epsilon \Pr(\mathcal{K}(D_2) \in S) +
\delta
\end{equation}
\end{definition}

If $\delta=0$, $\mathcal{K}$ is said to satisfy $\epsilon$-differential privacy.

To achieve DP, noise is added to the algorithm's output. This noise is proportional to the sensitivity of the output, where sensitivity measures the maximum change of the output due to the inclusion of a single data instance.

Two popular mechanisms for achieving DP are the Laplacian and Gaussian mechanisms. Gaussian is defined by
\begin{equation}\label{eq:gaussian}
M(D)\triangleq f(D) + N (0, S^2_f \sigma^2),
\end{equation}
where $N (0, S^2_f \sigma^2)$ is the normal distribution
with mean 0 and standard deviation $S_f \sigma$. A single application
of the Gaussian mechanism to function $f$ of sensitivity
$S_f$ satisfies $(\epsilon, \delta)$-differential privacy if $\delta \geq \frac{5}{4}exp(-(\sigma\epsilon)^2/2)$ and $\epsilon < 1$~\cite{dwork2014algorithmic}.

To achieve $\epsilon$-differential privacy, the Laplace mechanism may be used in the same manner by substituting
$N(0, S^2_f \sigma^2)$
with random variables drawn from
$Lap(S_f/\epsilon)$
~\cite{dwork2014algorithmic}.

When an algorithm requires multiple additive noise mechanisms, the evaluation of the privacy guarantee follows from the basic composition theorem
~\cite{dwork2006our,dwork2009differential} or from advanced composition theorems and their extensions~\cite{dwork2010boosting,dwork2016concentrated,kairouz2017composition,bun2016concentrated}.


\subsection{Threshold Homomorphic Encryption}

An additively homomorphic encryption scheme is one wherein the following property is guaranteed:
\begin{equation*}
    Enc(m_1) \circ Enc(m_2) = Enc(m_1 + m_2),
\end{equation*}
for some predefined function $\circ$. Such schemes are popular in privacy-preserving data analytics as untrusted parties can perform operations on encrypted values.

One such additive homomorphic scheme is the Paillier cryptosystem~\cite{paillier1999public}, a probabilistic encryption scheme based on computations in the group $\mathbb{Z}_{n^2}^{*}$, where $n$ is an RSA modulus. In~\cite{Damgard:2001:GSA:648118.746742} the authors extend this encryption scheme and propose a threshold variant. In the threshold variant, a set of participants is able to share the secret key such that no subset of the parties 
smaller than a pre-defined threshold is able to decrypt values.

\subsection{Privacy in Federated Learning}

In centralized learning environments a single party $P$ using a dataset $D$ executes some learning algorithm $f_M$ resulting in a model $M$ where $f_M(D) = M$. In this case $P$ has access to the complete dataset $D$. By contrast, in a federated learning environment, multiple parties $P_1, P_2, ..., P_n$, each have their own dataset $D_1, D_2, ..., D_n$, respectively. The goal is then to learn a model using all of the datasets.

We must consider two potential threats to data privacy in such an FL environment: (1) inference during the learning process and (2) inference over the outputs. \textit{Inference during the learning process} refers to any participant in the federation inferring information about another participant's private dataset given the data exchanged during the execution of $f_M$. \textit{Inference over the outputs} refers to the leakage of any participants' data from intermediate outputs as well as $M$.

We consider two types of inference attacks: insider and outsider. \textit{Insider attacks} include those launched by participants in the FL system, including both data holders as well as any third parties, while \textit{outsider attacks} include those launched both by eavesdroppers to the communication between participants and by users of the final predictive model when deployed as a service.

\subsubsection{Inference during the learning process}

Let us consider $f_M$ as the combination of computational operations and a set of queries $Q_1, Q_2, ..., Q_k$. That is, for each step $s$ in $f_M$ requiring knowledge of the parties' data there is a query $Q_s$. In the execution of $f_M$ each party $P_i$ must respond to each such query $Q_s$ 
with appropriate information on $D_i$. The types of queries are highly dependent on $f_M$. For example, to build a decision tree, a query may request the number of instances in $D_i$ matching a certain criteria. In contrast, to train an SVM or neural network a query would request model parameters after a certain number of training iterations.
Any privacy-preserving FL system must account for the risk of inference over the responses to these queries.

Privacy-preserving ML approaches addressing this risk often do so by using secure multiparty computation (SMC). Generally, SMC protocols allow $n$ parties to obtain the output of a function over their $n$ inputs while preventing knowledge of anything other than this output~~\cite{goldreich1998secure}. 
Unfortunately, approaches exclusively using secure multiparty computation remain vulnerable to inference over the output.
As the function output remains unchanged from function execution without privacy, the output can reveal
information about individual inputs. 
Therefore, we must also consider potential inference over outputs.
\subsubsection{Inference over the outputs}

This refers to intermediate outputs available to participants as well as the predictive model. Recent work shows that given only black-box access to the model through an ML as a service API, an attacker can still make training data inferences~\cite{shokri2017membership}. An FL system should prevent such outsider attacks while also considering insiders. That is, participant $P_i$ should not be able to infer information about $D_j$ when $i \neq j$ as shown in~\cite{nasr2018comprehensive}.

Solutions 
addressing privacy of output often make use of the DP framework discussed in Preliminaries. As a mechanism satisfying differential privacy guarantees that if an individual contained in a given dataset is removed, no outputs would become significantly more or less likely~~\cite{dwork2008differential}, 
a learning algorithm $f_M$ which is theoretically proven to be $\epsilon$-differentially private
is guaranteed to have a certain privacy of output quantified by the $\epsilon$ privacy parameter.

In the federated learning setting it is important to note that the definition of neighboring databases is consistent with the usual DP definition -- that is, privacy is provided at the individual record level, not the party level (which may represent many individuals).

\section{An End-to-End Approach with Trust}

\subsection{Threat Model}

We propose a system wherein $n$ data parties use an ML service for FL. We refer to this service as the \textit{aggregator}. Our system is designed to withstand three potential adversaries: 
(1) the aggregator, (2) the data parties, and (3) outsiders.
\subsubsection{Honest-But-Curious Aggregator}

The honest-but-curious or semi-honest adversarial model is commonly used in the field of SMC since its introduction in~\cite{beaver1991foundations} 
and application to data mining in~\cite{lindell2000privacy}. Honest-but-curious adversaries follow the protocol instructions correctly but will try to learn additional information. Therefore, the aggregator \textit{will not} vary from the predetermined ML algorithm but \textit{will} attempt to infer private information using all data received throughout the protocol execution. 

\subsubsection{Colluding Parties}

Our work also considers the threat of collusion among parties, including the aggregator
, through the trust parameter $t$ which 
is the minimum number of non-colluding parties. 
Additionally, in contrast to the aggregator, 
we consider scenarios in which parties in $\mathcal{P}$ may deviate from the protocol execution to achieve additional information on data held by honest parties.

\subsubsection{Outsiders}

We also consider potential attacks from adversaries outside of the system. Our work ensures that any adversary monitoring communications during training cannot infer the private data of the participants. We also consider users of the final model as potential adversaries. A predictive model output from our system may therefore be deployed as a service, remaining resilient to inference against adversaries who may be users of the service.

We now detail the assumptions made in our system to more concretely formulate our threat model.

\subsubsection{Communication}

We assume secure channels between each party and the aggregator. This allows the aggregator to authenticate incoming messages and 
prevents an adversary, whether they be an outsider or malicious data party, 
from injecting their own responses.

\subsubsection{System set up}

We additionally make use of the \textit{threshold variant of the Paillier encryption scheme} from~\cite{Damgard:2001:GSA:648118.746742} assuming secure key distribution. It is sufficient within our system to say that semantic security of encrypted communication is equivalent to the decisional composite residuosity assumption. For further discussion we direct the reader to~\cite{Damgard:2001:GSA:648118.746742}.
Our use of the threshold variant of the Paillier system ensures that any set of $n-t$ or fewer parties cannot decrypt ciphertexts. Within the context of our FL system, this ensures the privacy of individual messages sent to the aggregator.

\subsection{Proposed Approach}

We propose an FL system that addresses risk of inference during the learning process, risk of inference over the outputs, \textit{and} trust. We combine methods from SMC and DP to develop protocols that guarantee privacy without sacrificing accuracy.

We consider the following scenario. There exists a set of $n$ parties $\mathcal{P} = P_1, P_2, ..., P_n$, a set of disjoint datasets $D_1, D_2, ..., D_n$ belonging to the respective parties and adhering to the same structure, and an aggregator $\mathcal{A}$. Our system takes as additional input three parameters: $f_M, \epsilon,$ and $t$. $f_M$ specifies the training algorithm, $\epsilon$ is the privacy guarantee against inference, and $t$ specifies the minimum number of honest, non-colluding parties.

\begin{figure}
    \vspace{-0.5\baselineskip}
    \centering
    \includegraphics[scale=0.50]{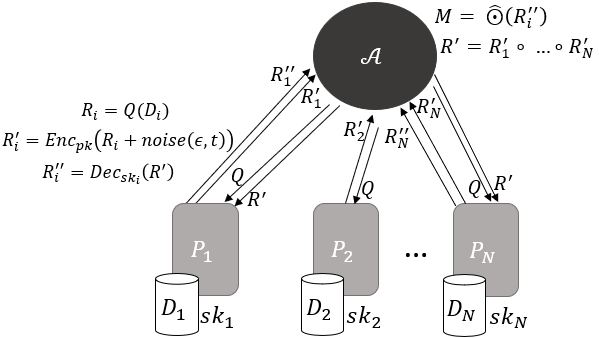}
    \vspace{-\baselineskip}
    \caption{System Overview}
    \label{fig:approach}
\end{figure}

The aggregator $\mathcal{A}$ runs the learning algorithm $f_M$ consisting of $k$ or fewer linear queries $Q_1, Q_2, ..., Q_k$, each requiring information from the $n$ datasets. This information may include model parameters after some local learning on each dataset or may be more traditionally queried information such as how many individuals match a set of criterion. For each algorithm deployed into our system, any step $s$ requiring such information reflective of some analysis on the private, local datasets must be represented by a corresponding query $Q_s$. Figure~\ref{fig:approach} shows an outline of a step $s$ in $f_M$. Using secure channels between $\mathcal{A}$ and each of the parties, the aggregator will send query $Q_s$.
Each party will then calculate a response using their respective datasets.
The participant will use a differential privacy mechanism that depends on the algorithm $f_M$ to add the appropriate amount of noise according to the privacy budget allocated to that step, the sensitivity of $Q_s$, and the level of trust in the system.
The noisy response is then encrypted using the threshold variant of the Paillier cryptosystem
and sent to $\mathcal{A}$. 
Homomorphic properties then allow $\mathcal{A}$ to aggregate the individual responses. $\mathcal{A}$ subsequently queries at least $n-t+1$ data parties to decrypt the aggregate value and updates the model $M$.
At the conclusion of $f_M$, the model $M$ is exposed to all participants.
This process 
is outlined in Algorithm~\ref{alg:main_algo}.

We consider trust with respect to collusion in two steps: (1) in the addition of noise and (2) in the threshold setting of the encryption scheme. 
The more participants colluding, the more knowledge which is available to infer the data of an honest participant. Therefore, the noise introduced by an honest participant must account for collusion. 
The use of homomorphic encryption however allows for significant increases in accuracy (over local privacy approaches). We now detail this strategy for FL.

\subsection{Reducing Noise with SMC}

A key component to our system is the ability to reduce noise by leveraging the SMC framework 
while considering a customizable trust parameter.

Specifically, 
let $\sigma_s$ and $S_s$ respectively be the noise parameter and sensitivity to step $Q_s$ allocated a budget $\epsilon_s$ in the learning algorithm $f_M$. 
In a traditional application of differential privacy to federated learning, each party will use the Gaussian mechanism to add $N(0, S^2_s \sigma_s^2)$ noise to their response $r_{i,s}$ when queried by $\mathcal{A}$ at step $Q_s$. This guarantees the privacy of each $r_{i,s}$.

If, however, each $r_{i,s}$ is encrypted using the scheme proposed in~\cite{Damgard:2001:GSA:648118.746742} with a threshold setting of $\bar{t} = n - t + 1$, the noise may be reduced by a factor of $t-1$. Rather than returning $Q_s(D_i) + N(0, S^2_s \sigma^2_s)$, each party may return $Enc(Q_s(D_i) + N(0, S^2_s \frac{\sigma^2_s}{t-1}))$.

Note that when $\mathcal{A}$ aggregates these responses the value that is eventually decrypted and exposed will be $\sum_{i=1}^{n} Q_s(P_i) + Y_i$ where each $Y_i$ is drawn from the Gaussian distribution with standard deviation $S_s \frac{\sigma_s}{\sqrt{t-1}}$. This is equivalent to $N(0, S^2_s \frac{n \sigma^2_s}{t-1}) \sum_{i=1}^n Q_s(D_i)$. Since we know that $t-1 < n$, the noise included in the decrypted value is strictly greater than that required to satisfy DP. Additionally, the encryption scheme guarantees that the maximum number of colluders, $\bar{t}$, cannot decrypt values of honest parties.


Given this approach, we are able to maintain the customizable nature of our system with the trust parameter $t$ and the formal privacy guarantees of the DP framework while decreasing the amount of noise \textit{for each query response} leading to more accurate ML models.

\newlength{\textfloatsepsave} \setlength{\textfloatsepsave}{\textfloatsep} \setlength{\textfloatsep}{0.25\baselineskip}
\begin{algorithm}[t]
\begin{footnotesize}
\caption{Private Federated Learning}\label{alg:main_algo}
\begin{algorithmic}
\STATE \textbf{Input}: ML algorithm $f_M$; set of data parties $\mathcal{P}$ of size $N$, with each $P_i \in \mathcal{P}$ holding a private dataset $D_i$ and a portion of the secret key $sk_i$; minimum number of honest, non-colluding parties $t$; privacy guarantee $\epsilon$
\STATE $\bar{t} = n - t + 1$
\FOREACH {$Q_s \in f_M$}
    \FOREACH{$P_i \in \mathcal{P}$}
        \STATE $\mathcal{A}$ asynchronously queries $P_i$ with $Q_s$
        \STATE $P_i$ sends $r_{i,s} = Enc_{pk}(Q_s(D_i) + noise(\epsilon,t))$
    \ENDFOR
    \STATE $\mathcal{A}$ aggregates $Enc_{pk}(r_s) \leftarrow r_{1,s} \circ r_{2,s} \circ ... \circ r_{N,s}$
    \STATE $\mathcal{A}$ selects $\mathcal{P}_{dec} \subseteq \mathcal{P} $ such that $\vert \mathcal{P}_{dec} \vert = \bar{t}$
    \FOREACH{$P_i \in \mathcal{P}_{dec}$}
        \STATE $\mathcal{A}$ asynchronously queries $P_i$ with $Enc_{pk}(r_s)$
        \STATE $\mathcal{A}$ receives partial decryption of $r_s$ from $P_i$ using $sk_i$
    \ENDFOR
    \STATE $\mathcal{A}$ computes $r_s$ from partial decryptions
    \STATE $\mathcal{A}$ updates $M$ with $r_s$
\ENDFOR
\STATE \textbf{return} $M$
\end{algorithmic}
\end{footnotesize}
\end{algorithm}

\section{Experimental Evaluation}

In this section we empirically demonstrate how to apply our approach to train three distinct learning models: decision trees (DT), convolutional neural networks (CNN) and linear Support Vector Machines (SVM). We additionally provide analysis on the impact of certain settings on the performance of our approach.

\subsection{Decision Trees}

We first consider DT learning using the ID3 algorithm.
In this scenario, each dataset $D_i$ owned by some $P_i \in \mathcal{P}$ contains a set of instances described by the same set of categorical features $\mathcal{F}$ and a class attribute $C$. The aggregator initializes the DT model with a root node. Then, the feature $F \in \mathcal{F}$ that maximizes information gain is chosen based on counts queried from each party in $\mathcal{P}$ and child nodes are generated for each possible value of $F$. The feature $F$ is then removed from $\mathcal{F}$. This process continues recursively for each child node until either (a) there are no more features in $\mathcal{F}$, (b) a pre-determined max-depth is reached, or (c) responses are too noisy to be deemed meaningful.
This process is specifically detailed as algorithmic pseudocode in Section~\ref{subsec:app_dt}.

There are two types of participant queries in private, federated DT learning
: \textit{counts} and \textit{class\_counts}. For executing these queries $\mathcal{A}$ first divides the entire privacy budget $\epsilon$ equally between each layer of the tree. 
According to the composition property of differential privacy, because different nodes within the same layer are evaluated on disjoint subsets of the datasets, they do not accumulate privacy loss and therefore the budget allocated to a single layer is not divided further. Within each node, half of the budget ($\epsilon_1$) is allocated to determining total counts and half is allocated to either class counts (done at the leaf nodes) or evaluating attributes (done at internal nodes). For internal nodes, each feature is evaluated for potential splitting against the same dataset. The budget allocated to evaluating attributes must therefore be divided amongst each feature ($\epsilon_2$). 
In all experiments the max depth is set to $d = \frac{|\mathcal{F}|}{2}$.

\paragraph{Dataset}
We conduct a number of experiments using the Nursery dataset from the UCI Machine Learning Repository~\cite{Dua:2017}. This dataset contains 8 categorical attributes about 12,960 nursery school applications. 
The target attribute has five distinct classes with the following distribution: $33.333 \%, 0.015 \%, 2.531 \%, 32.917 \%, 31.204 \%$.

\paragraph{Comparison Methods}
To put model performance 
into context, we compare with two different random baselines and two current FL approaches. Random baselines enable us to characterize when a particular approach is no longer learning meaningful information while the FL approaches visualize relative performance cost. 
\begin{enumerate}
    \item Uniform Guess. In this approach, class predictions are randomly sampled with a $\frac{1}{|C|}$ chance for each class.
    \item Random Guess. Random guess improves upon Uniform Guess with consideration of class value distribution in the training data. At test time, each prediction is sampled from the set of training class labels.
    \item Local DP. In the local approach, parties add noise to protect the privacy of their own data in isolation. The amount of noise necessary to provide $\epsilon$-differential privacy to each dataset is defined in~\cite{blum2005practical}.
    \item No Privacy. This is the result of running the distributed learning algorithm without any 
    privacy guarantee.
\end{enumerate}

\subsubsection{\textbf{Variation in Settings}}\label{subsubsec:variation_settings}

We now look at how different settings impact results.

\paragraph{Privacy Budget}
We first look at the impact of the privacy budget on performance in our system. To isolate the impact of the privacy budget we set the number of parties, $|\mathcal{P}|$, to 10 and assume no collusion. 
We consider 
budget values between 0.05 and 2.0. Recall from Preliminaries that for a mechanism to be $\epsilon$-differentially private the amount of noise added will be inversely proportional to value of $\epsilon$. In other words, the smaller the $\epsilon$ value, the smaller the privacy budget, and the more noise added to each query.

\setlength{\textfloatsep}{\textfloatsepsave}
\begin{figure}[t]
    \vspace{-0.5\baselineskip}
    \centering
    \includegraphics[width=0.75\columnwidth]{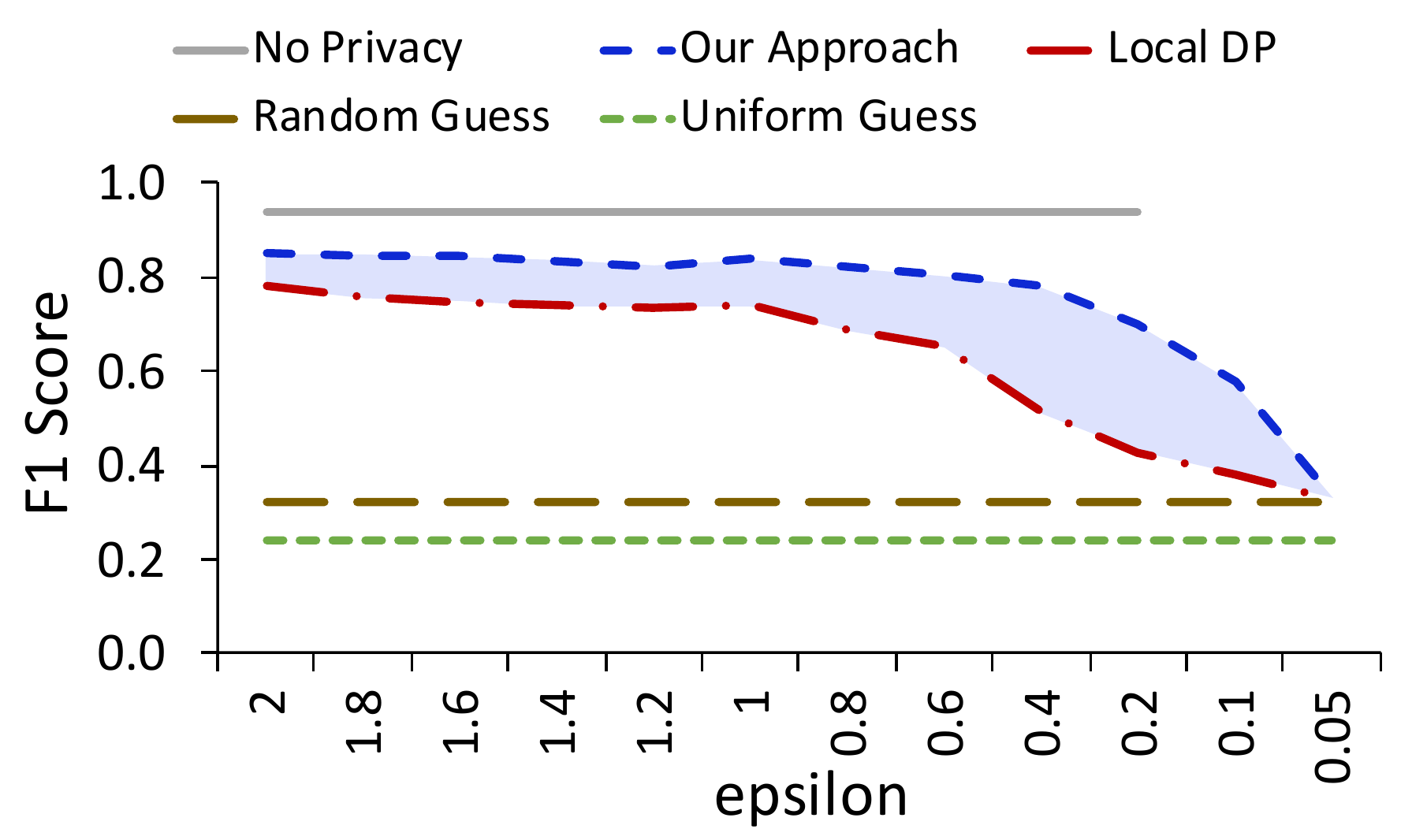}
    \vspace{-\baselineskip}
    \caption{Effect of privacy budgets on the overall F1-score for Decision Trees}
    \label{fig:dt_nursery}
\end{figure}

We can see in Figure \ref{fig:dt_nursery} that our approach maintains an F1-score above 0.8 for privacy budgets as small as 0.4. Once the budget dips below 0.4 we see the noise begins to overwhelm the information being provided which can have one of two outcomes: (1) learning pre-maturely haults or (2) learning become inaccurate. This results in degraded performance as the budget decreases, which is expected. It is clear 
that our approach maintains improved performance over the local DP approach for all budgets (until both approaches converge to the random guessing baseline). Particularly as the budget decreases from 1.0 to 0.4 we see our approach maintaining better resilience to the decrease in the privacy budget. 

\paragraph{Number of Parties}
Another important consideration for FL systems is the ability to maintain accuracy in highly distributed scenarios. That is, when many parties, each with a small amount of data, such as in an IoT scenario, are contributing to the learning. 

\begin{figure}[t]
    \vspace{-0.5\baselineskip}
    \centering
    \includegraphics[width=0.75\columnwidth]{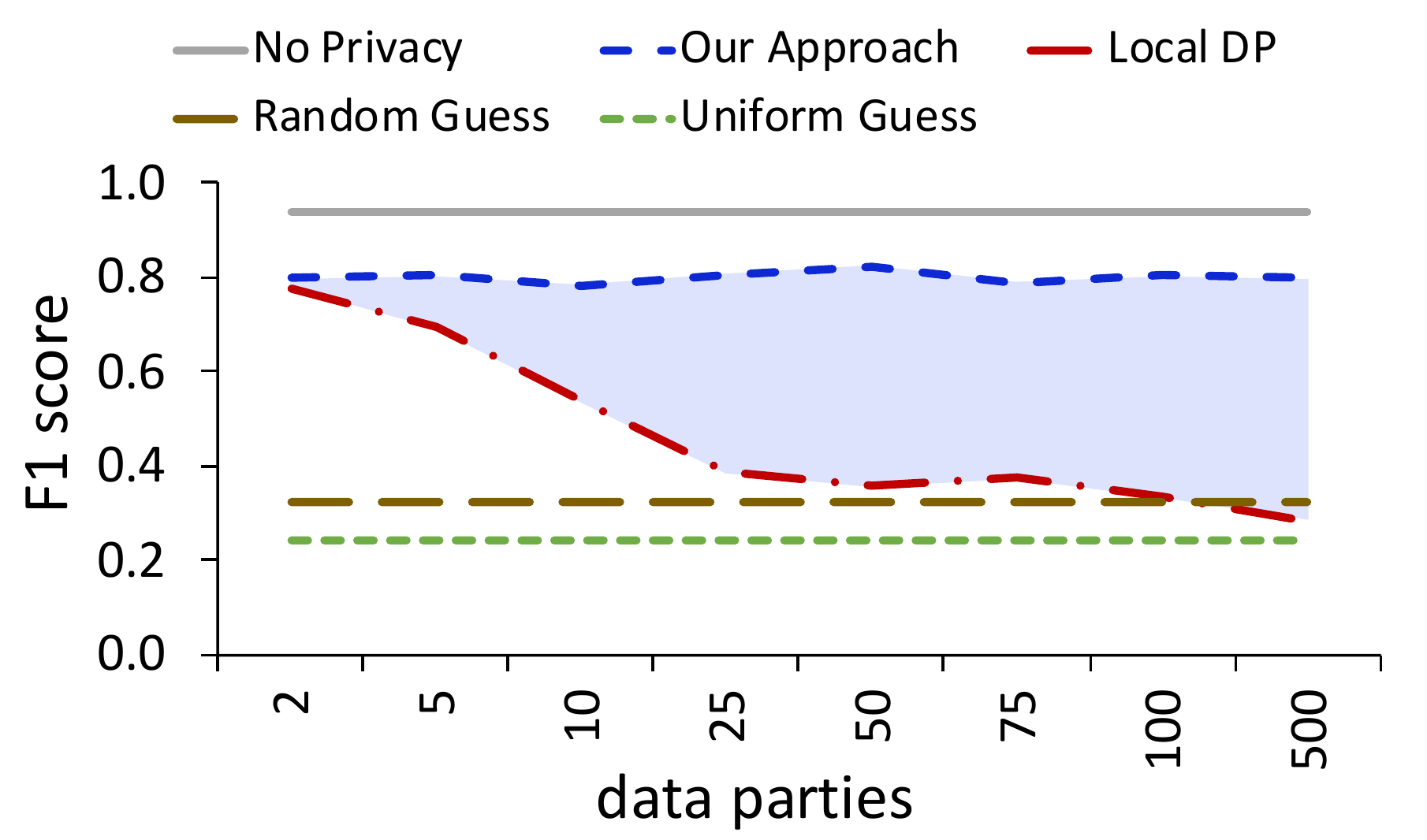}
    \vspace{-\baselineskip}
    \caption{Effect of increasing number of parties on the overall F1-score for Decision Trees}
    \label{fig:num_parties}
\end{figure}

\begin{figure}[h]
    \vspace{-0.5\baselineskip}
    \centering
    \includegraphics[width=0.75\columnwidth]{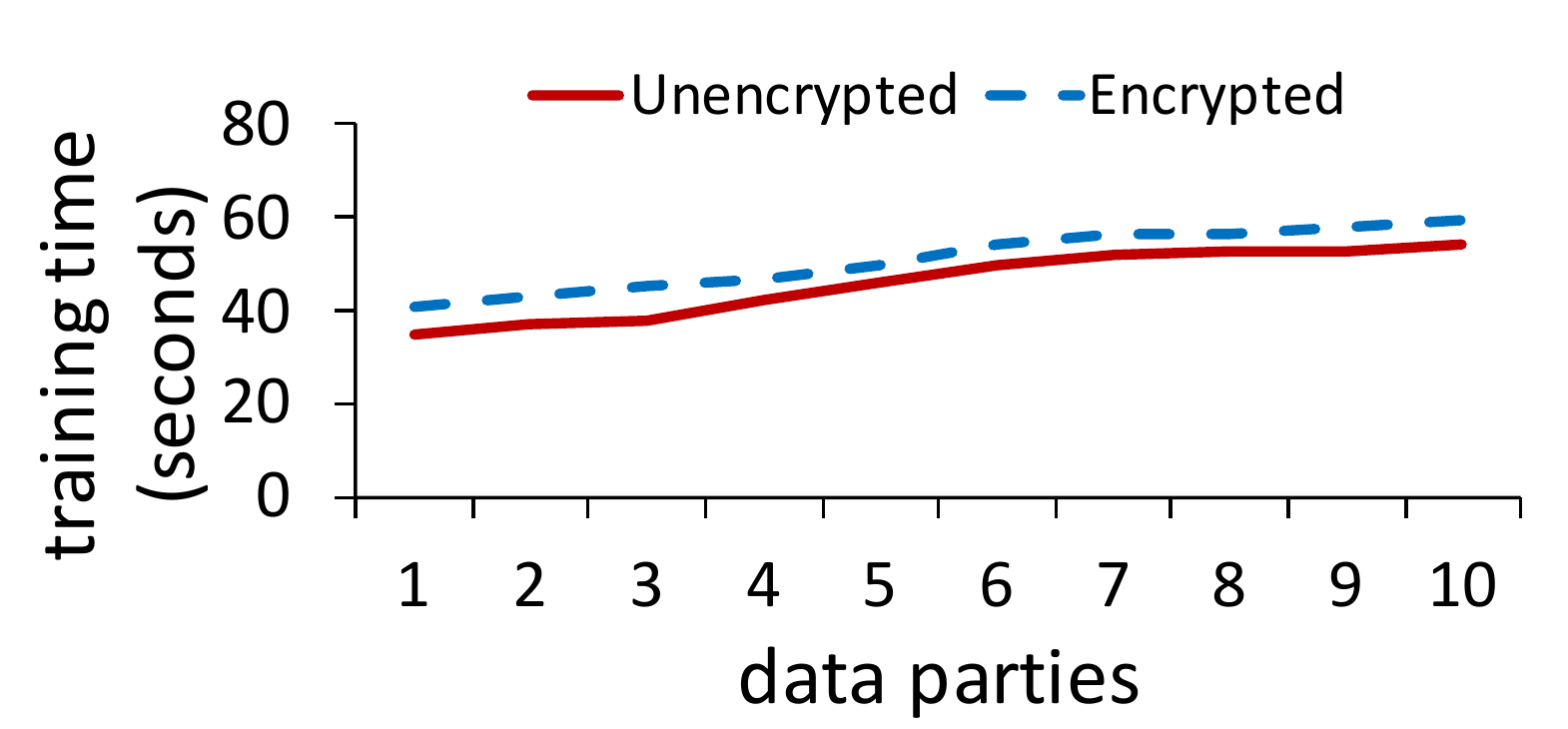}
    \vspace{-\baselineskip}
    \caption{Decision Tree Training Time with Encryption}
    \label{fig:train_time}
\end{figure}

In Figures \ref{fig:num_parties} and \ref{fig:train_time} we show the impact that $|\mathcal{P}|$ has on performance. The results are for a fixed overall privacy budget of 0.5 and assume no collusion. For each experiment, the overall dataset was divided into $|\mathcal{P}|$ equal sized partitions. 

The results in Figure \ref{fig:num_parties} demonstrate the viability of our system for FL in highly distributed environments 
while highlighting the shortcomings of the local DP approach. 
As $|\mathcal{P}|$ increases, the noise in the local DP approach increases proportionally while 
our approach 
maintains consistent accuracy. 
We can see that with as few as 25 parties, the local DP results begin to approach the baseline and even dip below random guessing by 100 participants.


\begin{figure}
    \vspace{-0.5\baselineskip}
    \centering
    \includegraphics[width=0.75\columnwidth]{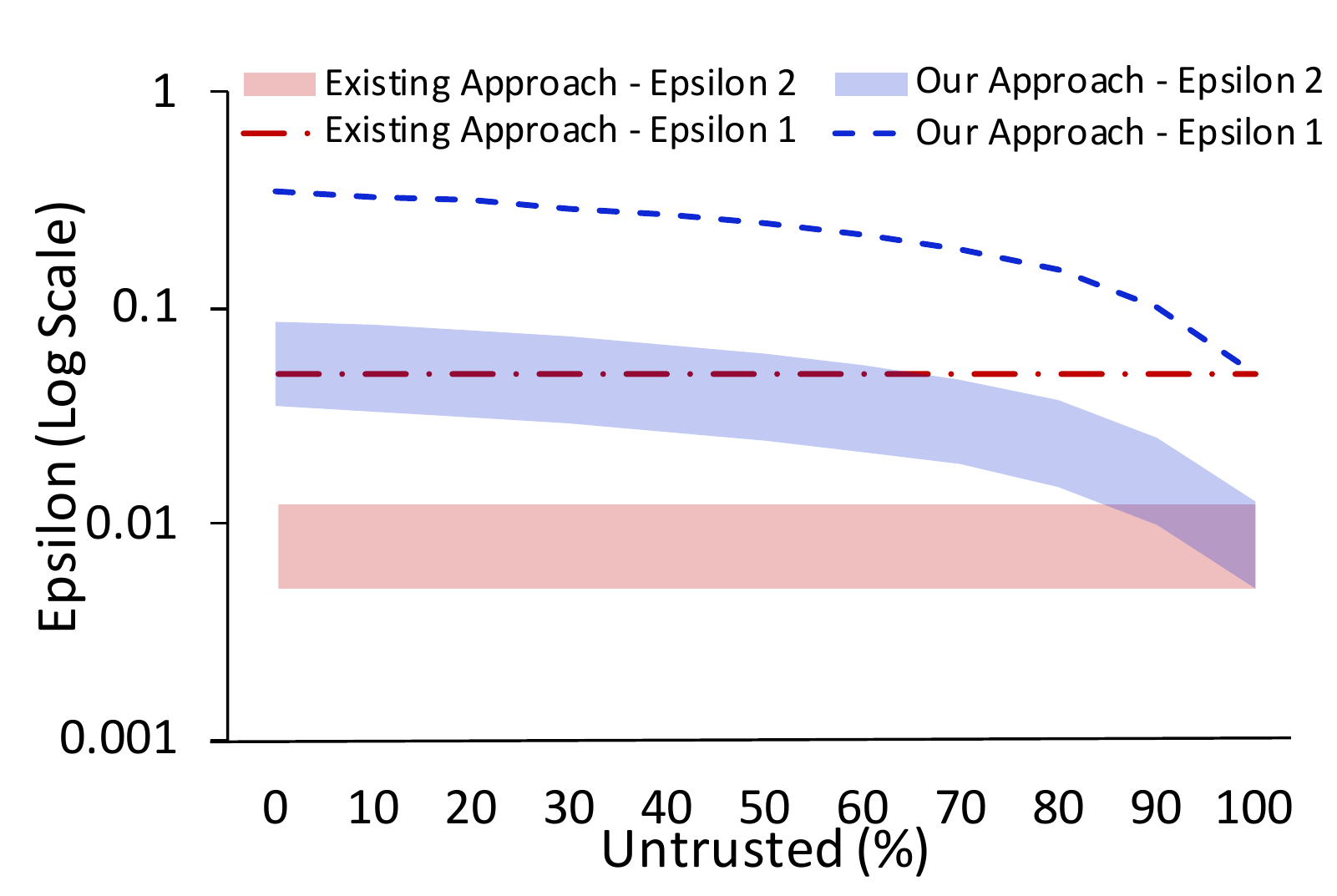}
    \vspace{-\baselineskip}
    \caption{Query Epsilons in Decision Tree Training with Varying Rate of Trust (50 parties). Epsilon 1 is defined as the privacy budget for count queries while Epsilon 2 is used for class counts.
    }
    \label{fig:trust}
\end{figure}

Another consideration relative to the scaling of participants is the overhead of encryption. Figure \ref{fig:train_time} highlights the scalability of our system, showing the impact that encryption has on overall training time in our system as the number of parties increases from 1 to 10. While the entire system experiences a steady increase in cost as the number of participants increases, the impact of the encryption remains consistent. Because our system is designed for a distributed scenario, the interactions with the aggregator are done in parallel and therefore the overhead of encryption remains constant as the number of parties increases. 

\paragraph{Trust}
An important part of our system is the trust parameter. While the definition of a neighboring database within the context of the differential privacy framework considers privacy at the record level, the trust model for adversarial knowledge is considered within the context of the entire system. The trust parameter therefore represents the degree of adversarial knowledge by capturing the maximum number of colluding parties which the system may tolerate. Figure \ref{fig:trust} demonstrates how the $\epsilon$ values used for both count and distribution queries in private, federated DT learning are impacted by the trust parameter setting when $|\mathcal{P}| = 50$.

In the worst case scenario where a party $P_i \in \mathcal{P}$ assumes that all other $P_j \in \mathcal{P}, i \neq j$ are colluding, our approach converges with existing local DP approaches. 
In all other scenarios the query $\epsilon$ values will be increased in our system leading to more accurate outcomes. Additionally, we believe the aforementioned scenario of no trust is unlikely to exist in real world instances. Let us consider smart phone users as an IoT example. 
Collusion of all but one party is impractical not only due to scale but also since such a system is likely the be running without many users even knowing. Additionally, on a smaller scale, if there is a set of five parties in the system and one party is concerned that the other four are all colluding, there is no reason for the honest party to continue to participate. We therefore believe that realistic scenarios of FL will see accuracy gains when deploying our system.

\subsection{Convolutional Neural Networks}

We additionally demonstrate how to use our method to train a distributed differentially private CNN. In our approach, similarly to centrally trained CNNs, each party is sent a model with the same initial structure and randomly initialized parameters. Each party will then conduct one full epoch of learning locally. At the conclusion of each batch, Gaussian noise is introduced according to the norm clipping value $c$ and the privacy parameter $\sigma$. Norm clipping allows us to put a bound on the sensitivity of the gradient update.
We use the same privacy 
strategy used in the centralized training approach 
presented in~\cite{abadi2016deep}.
Once an entire epoch, or $\frac{1}{b}$ batches where $b =$ batch rate, has completed the final parameters are sent back to $\mathcal{A}$. $\mathcal{A}$ then averages the parameters and sends back an updated model for another epoch of learning. After a pre-determined $E$ number of epochs, 
the final model $M$ is output. This process for the aggregator and data parties are specifically detailed as algorithmic pseudocode in Section~\ref{subsec:app_nn}.

\setlength{\textfloatsepsave}{\textfloatsep} \setlength{\textfloatsep}{0.25\baselineskip}
Within our private, federated NN learning system, if $\sigma = \sfrac{\sqrt{2\cdot\log\frac{1.25}{\delta}}}{\epsilon}$ then by \cite{dwork2014algorithmic} 
our approach is $(\epsilon, \delta)$-differentially private with respect to each randomly sampled batch. 
Using the moments accountant in~\cite{abadi2016deep}, 
our approach is $(O(b\epsilon \sqrt{\smash[b]{\sfrac{E}{b}}}),\delta)$-DP overall.


\paragraph{Dataset and Model Structure}
For our CNN experiments we use the publicly available MNIST dataset. This includes 60,000 training instances of handwritten digits and 10,000 testing instances. Each example is a 28x28 grey-scale image of a digit between 0 and 9~\cite{lecun1998gradient}.


We use a model structure similar to that in~\cite{abadi2016deep}. Our model is a feedforward neural network with 2 internal layers of ReLU units and a softmax layer of 10 classes with cross-entropy loss. The first layer contains 60 units 
and the second layer contains 1000. We set the norm clipping to 4.0, learning rate to 0.1 and batch rate to 0.01. We use Keras 
with a Tensorflow 
backend.

\paragraph{Comparison Methods}
To the best of our knowledge, this paper presents the first approach to accurately train a neural network in a private federated fashion without reliance on any public or non-protected data.
We therefore compare our approach with the following baselines:
\begin{enumerate}
    \item Central Data Holder, No Privacy. In this approach all the data is centrally held by one party and no privacy is considered in the learning process.
    \item Central Data Holder, With Privacy. While all the data is still centrally held by one entity, this data holder now conducts privacy-preserving learning. This is representative of the scenario in~\cite{abadi2016deep}.
    \item Distributed, No Privacy. In this approach the data is distributed to multiple parties, but the parties do not add noise during the learning process.
    \item Local DP. Parties add noise to protect the privacy of their own data in isolation, adapting from~\cite{abadi2016deep} and~\cite{shokri2015privacy}.
\end{enumerate}

\begin{figure}[t]
    \vspace{-0.5\baselineskip}
    \centering
    \includegraphics[width=0.75\columnwidth]{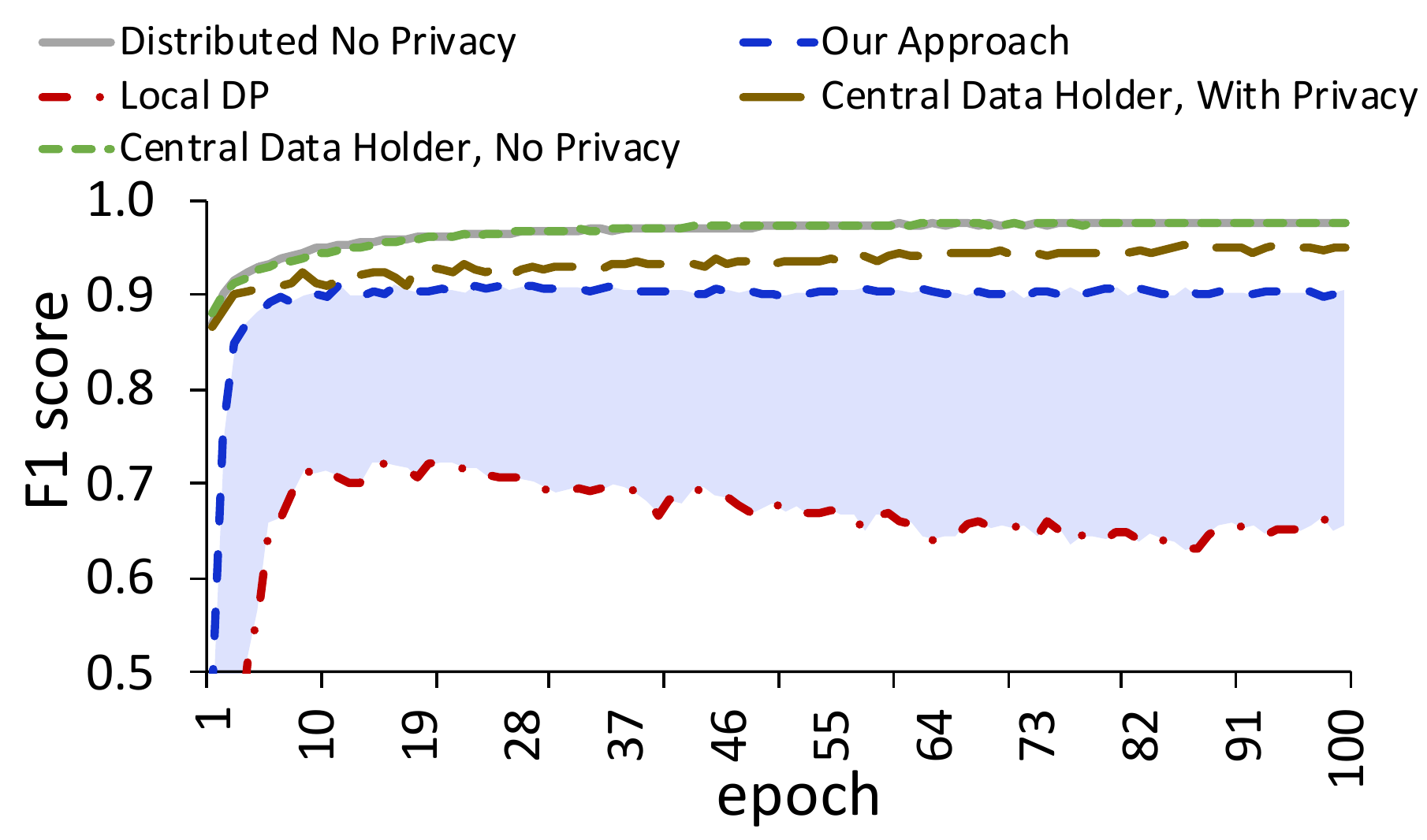}
    \vspace{-\baselineskip}
    \caption{Convolutional Neural Network Training with MNIST Data (10 parties and $\sigma$ = 8, ($\epsilon$, $\delta$) = (0.5, $10^{-5}$))}
    \label{fig:nn_mnist}
\end{figure}

Figure \ref{fig:nn_mnist} shows results with 
10 parties conducting 100 epochs of training with the privacy parameter $\sigma$ set to 8.0, the ``large noise'' setting in~\cite{abadi2016deep}. Note that Central Data Holder, No Privacy and Distributed Data, No Privacy achieve similar results and thus overlap. Our model is able to achieve an F1-score in this setting of 0.9. While this is lower than the central data holder setting where an F1-score of approximately 0.95 is achieved, our approach again significantly out-performs the local approach which only reaches 0.723. Additionally, we see a drop off in the performance of the local approach early on as updates become overwhelmed by noise.

\setlength{\textfloatsep}{\textfloatsepsave}

We additionally experiment with $\sigma = 4$ and $\sigma = 2$ as was done in~\cite{abadi2016deep}. When $\sigma = 4$ (($\epsilon$, $\delta$) = (2, $10^{-5}$)) the central data holder with privacy is able to reach an F1 score of 0.960, the local approach reaches 0.864, and our approach results in an F1-score of 0.957. When $\sigma = 2$ (($\epsilon$, $\delta$) = (8, $10^{-5}$)) those scores become 0.973, 0.937, and 0.963 respectively. We can see that our approach therefore demonstrates the most gain with larger $\sigma$ values which translates to tighter privacy guarantees. 

\begin{figure}[t]
    \vspace{-0.5\baselineskip}
    \centering
    \includegraphics[width=0.75\columnwidth]{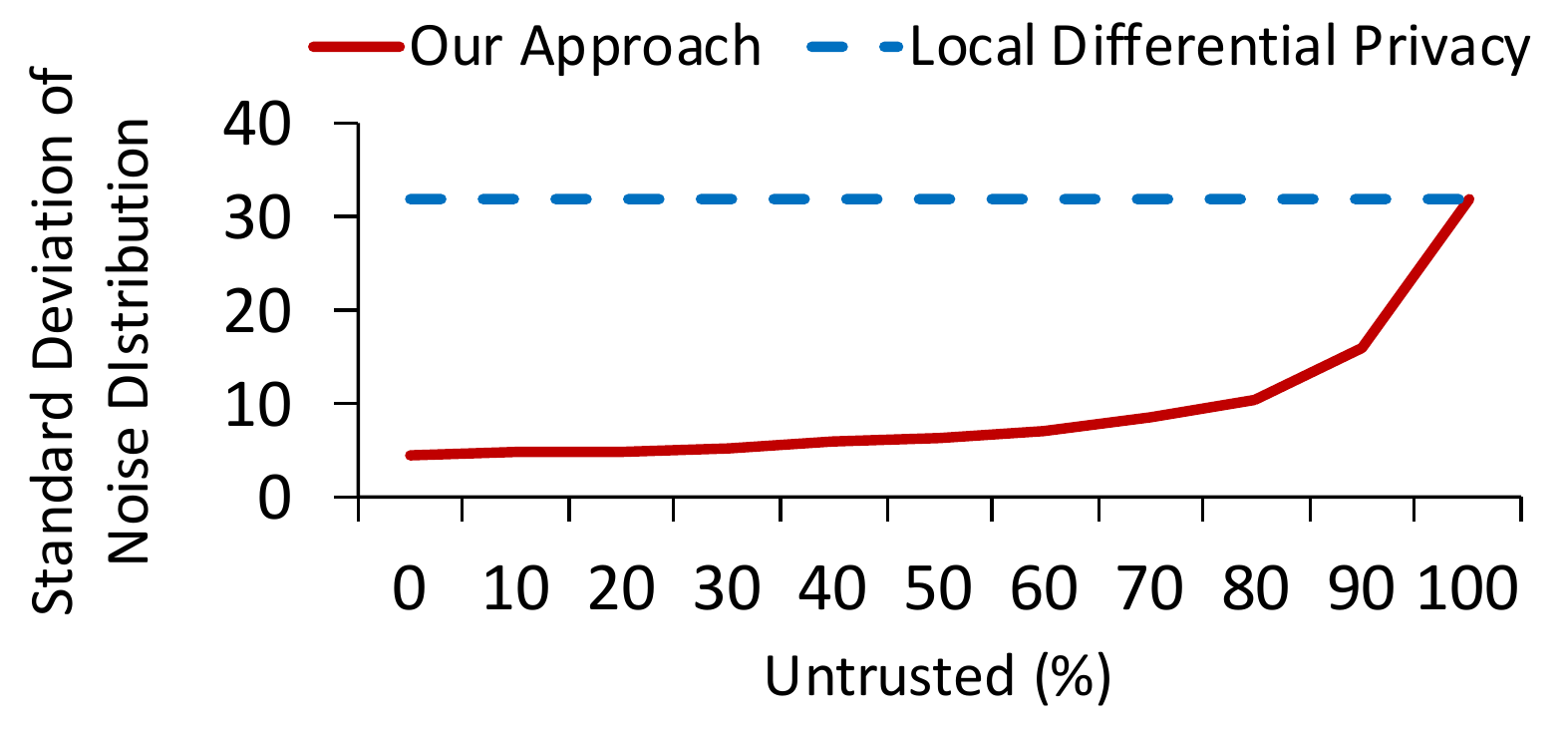}
    \vspace{-\baselineskip}
    \caption{Degree of Noise in Convolutional Neural Network Training with Varying Rate of Trust}
    \label{fig:trust_NN}
\end{figure}

Figure \ref{fig:trust_NN} again shows how the standard deviation of noise is significantly decreased in our system for most scenarios.

Our experiments demonstrate that the encryption time for one parameter at a party $P_i$ takes approximately 0.001095 sec while decryption between $\mathcal{A}$ and $\mathcal{P}_{dec}$ takes 0.007112 sec. While each parameter requires encryption and decryption, these processes can be done completely in parallel. Therefore overhead remains relatively constant as both $|\mathcal{P}|$ and the number of parameters increase.

Overall, we have demonstrated that our system provides significant accuracy gains for FL compared with local DP in plausible, real world scenarios and scales well. 

\subsection{Support Vector Machines (SVM)}

We also demonstrate and assess our approach when solving a classic
$\ell_2$-regularized binary linear SVM problem with hinge loss.

To train the linear SVM in a private distributed fashion, the aggregator distributes a model with the same weight vector $w$ to all parties.
Each party then runs a predefined number of epochs to learn locally.
To apply differential privacy in this setting, we first perform norm clipping on the feature vector $x$ to obtain a bound on the sensitivity of the gradient update.
Then, Gaussian noise is added to the gradient according to \cite{private-svm-gd}.
After each party completes its local training, the final noisy encrypted weights are sent back to the aggregator.
The aggregator averages the encrypted weights and sends back an updated model with a new weight vector for another epoch of learning.
Training ends after a predefined number of epochs. The detailed process is presented in Section~\ref{subsec:app_svm}.

\paragraph{Dataset}
We use the publicly available `gisette' dataset, which was used for NIPS 2003 Feature Selection Challenge \cite{libsvm}.
This dataset has $6,000$ training and $1,000$ testing samples with $5,000$ features.

\paragraph{Comparison Methods}
We contrast the performance of our approach against other ways to train the model.
\begin{enumerate}
    \item Central, No Privacy. Centralized training without privacy.
    \item Central DP. Centralized training with DP.
    \item Distributed, No Privacy. Model trained through federated learning without privacy.
    \item Local DP. In this approach each party adds enough noise independently to protect their data according to \cite{blum2005practical}.
\end{enumerate}

\setlength{\textfloatsep}{\textfloatsepsave}
\begin{figure}[t]
    \vspace{-0.5\baselineskip}
    \centering
    \includegraphics[width=0.75\columnwidth]{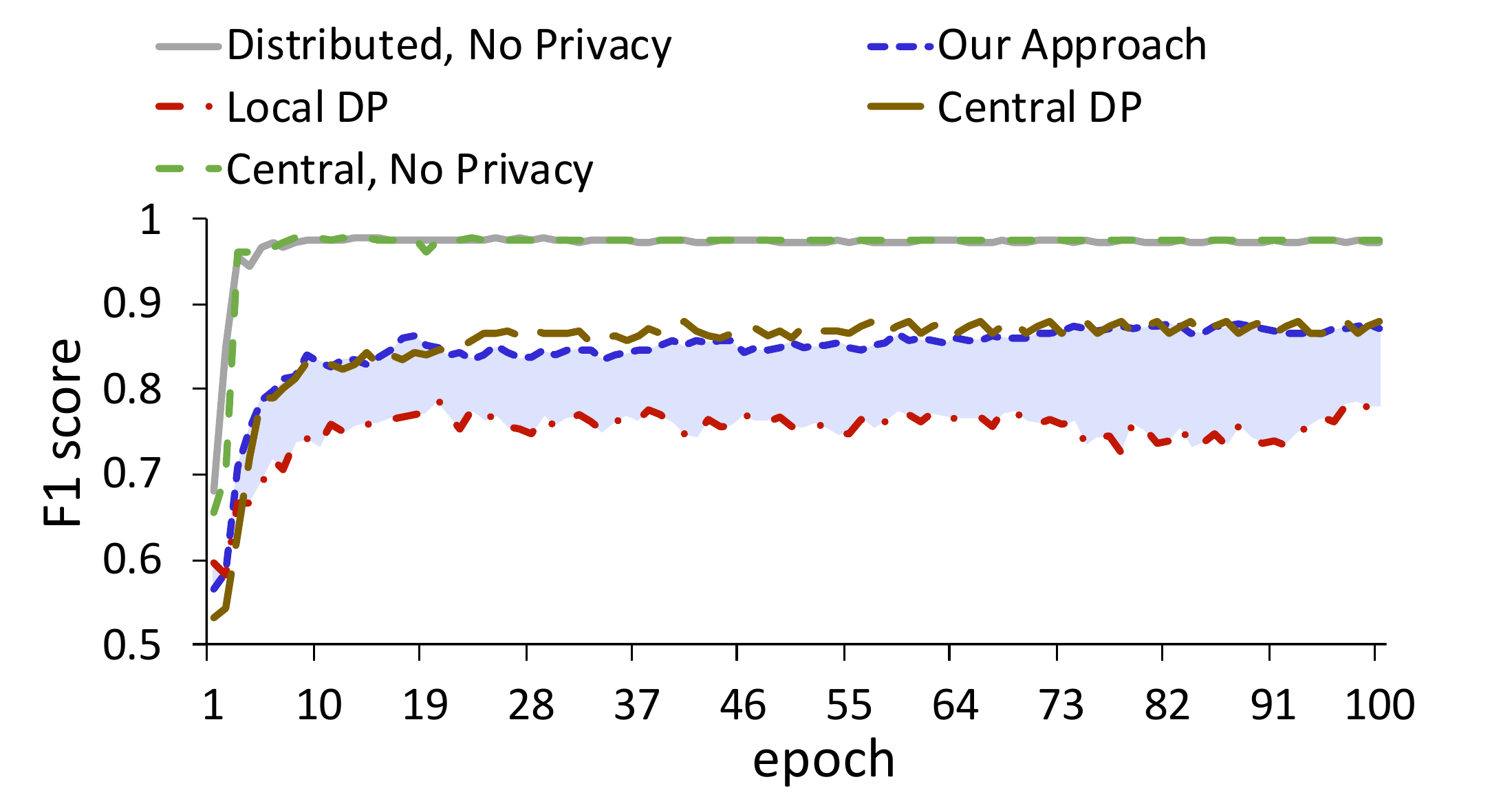}
    \vspace{-\baselineskip}
    \caption{Linear SVM Training(10 parties and ($\epsilon$, $\delta$) = (5, 0.0059))}
    \label{fig:svm}
\end{figure}

In these experiments, the learning rate was set to 0.01 for all settings.
We used 100 epochs for all approaches. Additionally, for
FL methods, each party runs 10 epochs locally.
We used 10 non-colluding parties.
Using $\sigma$=4, we report findings to achieve $(\epsilon, \delta)= (5, 0.0059)$ according to \cite{bun2016concentrated}.

Figure~\ref{fig:svm} shows F1-scores for the evaluated training methods.
Central, No Privacy and Distributed, No Privacy perform similarly with F1-scores around 0.99 after fewer than 10 epochs due to the balanced nature of the dataset. Among the privacy-preserving approaches, Central DP introduces the least noise and achieves the highest F1-score.
Among private FL methods, our approach achieves an F1-score over 0.87 which is almost equal to Central DP and significantly out-performs Local DP after 100 epochs.

We also evaluated our system in a lower trust setting with only half of the parties trusted as non-colluding. Our approach again out-performed Local DP. Specifically, after 100 epochs, our approach reached an F1-score of 0.85, while the Local DP achieves only 0.75.

These experimental results show that
our approach consistently out-performs state of the art methods to train different ML models in a private FL fashion.
We similarly showed that our approach consistently out-performs baselines such as random guessing while remaining reasonably close to non-private settings.

\section{System Implementation}

The development and deployment of new machine learning training algorithms to our system requires the training process be first broken down into a set of queries in which meaningful aggregation may be done via summation. Each query must then be analyzed for its privacy impact and and designated a portion of the overall privacy budget. Additionally, support must exist at each party for each type of query required by the private training algorithm. We will now provide implementation details for each of the model types evaluated, Decision Trees, Neural Networks, and Support Vector Machines, 
with additional discussion on the applicability of our framework to machine learning algorithms for other model types.

\subsection{Application to Private Decision Tree Training}\label{subsec:app_dt}

DT learning follows these steps: (1) determine the feature which best splits the training data, (2) split the training data into subsets according to the chosen feature, (3) repeat steps (1) and (2) for each subset. This is repeated until the subsets have reached a pre-determined level of uniformity with respect to the target variable.

\begin{algorithm}[ht]
\begin{footnotesize}
\caption{Private Decision Tree Learning}\label{alg:dt_algo_A}
\begin{algorithmic}
\STATE \textbf{Input}: Set of data parties $\mathcal{P}$; minimum number of honest, non-colluding parties $t$; privacy guarantee $\epsilon$; attribute set $\mathcal{F}$; class attribute $C$; max tree depth $d$; public key $pk$
\STATE $\bar{t} = n - t + 1$
\STATE $\epsilon_1 = \frac{\epsilon}{2(d+1)}$
\STATE Define current splits, $\mathcal{S} = \emptyset$, for root node
\STATE $M$ = BuildTree($\mathcal{S}$, $\mathcal{P}$, $t$, $\epsilon_1$, $\mathcal{F}$, $C$, $d$, $pk$)
\STATE \textbf{return} $M$
\STATE \textbf{procedure} \textsc{BuildTree}($\mathcal{S}$, $\mathcal{P}$, $t$, $\epsilon_1$, $\mathcal{F}$, $C$, $d$, $pk$)
\bindent
    \STATE $f = \text{max}_{F \in \mathcal{F}}|F|$
    \STATE Asynchronously query $\mathcal{P}$: \textit{counts}($\mathcal{S}, \epsilon_1$, $t$)
    \STATE $N = $ decrypted aggregate of noisy counts
    \IF{$\mathcal{F} = \emptyset$ or $d=0$ or $\frac{N}{f|C|} < \frac{\sqrt{2}}{\epsilon_1}$}
    \bindent
        \STATE Asynchronously query $\mathcal{P}$: \textit{class\_counts}($\mathcal{S}, \epsilon_1$, $t$)
        \STATE $N_c = $ vector of decrypted, noisy class counts
        \STATE \textbf{return} node labeled with $\argmax_c N_c$
    \eindent
    \ELSE
    \bindent
        \STATE $\epsilon_2 = \frac{\epsilon_1}{2|\mathcal{F}|}$
        \FOREACH{$F \in \mathcal{F}$}
        \bindent
            \FOREACH{$f_i \in F$}
            \bindent
                \STATE Update set of split values to send to child node: $S_i = S + \{F=f_i\}$
                \STATE Asynchronously query $\mathcal{P}$: \textit{counts}($\mathcal{S}_i$, $\epsilon_2$, $t$)
                \STATE and \textit{class\_counts}($\mathcal{S}_i$, $\epsilon_2$, $t$)
                \STATE ${N'}^F_i = $ aggregate of \textit{counts}
                \STATE ${N'}^F_{i,c} = $ element-wise aggregate of \textit{class\_counts}
                \STATE Recover $N^F_i$ from $\bar{t}$ partial decryptions of ${N'}^F_i$
                \STATE Recover $N^F_{i,c}$ from $\bar{t}$ partial decryptions of ${N'}^F_{i,c}$
            \eindent
            \ENDFOR
            \STATE $V_F = \sum_{i=1}^{|F|} \sum_{c=1}^{|C|} N_{i,c}^F \cdot \text{log} \frac{N^F_{i,c}}{N^F_i}$
        \eindent
        \ENDFOR
        \STATE $\bar{F} = \argmax_F V_F$
        \STATE Create root node $M$ with label $\bar{F}$
        \FOREACH{$f_i \in \bar{F}$}
        \bindent
            \STATE $S_i = S + \{F=f_i\}$
            \STATE $M_i = $ \textit{BuildTree}($S_i$, $\mathcal{P}$, $t$, $\epsilon_1$, $\mathcal{F}\setminus\bar{F}$, $C$, $d-1$, $pk$)
            \STATE Set $M_i$ as child of $M$ with edge $f_i$
        \eindent
        \ENDFOR
        \STATE \textbf{return} $M$
    \eindent
    \ENDIF
\eindent
\STATE \textbf{end procedure}
\end{algorithmic}
\end{footnotesize}
\end{algorithm}

To conduct private decision tree learning in our system we first address step (1): determining the best feature on which to split the data. We define the ``best'' feature as the feature which maximizes information gain. This is the same metric used in the ID3~\cite{quinlan1986induction}, C4.5~\cite{quinlan1993c4} and C5.0~\cite{quinlan2007c5} tree training algorithms. Information gain for a candidate feature $f$ quantifies the difference between the entropy of the current data with the weighted sum of the entropy values for each of the data subsets which would be generated if $f$ were to be chosen as the splitting feature. Entropy for a dataset (or subset) $D$ is computed via the following equation:
\begin{equation}
    Entropy(D) = \sum_{i=1}^{|C|} p_i \text{log}_2 p_i
\end{equation}
where $C$ is the set of potential class values and $p_i$ indicates the probability that a random instance in $D$ is of class $i$. Therefore, the selection of the ``best'' feature on which to switch can be chosen via determining class probabilities which in turn may be computed via counts. Queries to the parties from the aggregator are therefore counts and class counts, known to have a sensitivity of 1.

Given the ability to query parties for class counts the aggregator may then control the iterative learning process. To ensure this process is differentially private according to a pre-defined privacy budget, we follow the approach from~\cite{friedman2010data} to divide the budget for each iteration and set a fixed number of iterations rather than a purity test as a stopping condition. The algorithm will also stop if counts appear too small relative to the degree of noise to provide meaningful information. The resulting private algorithm deployed in our system is detailed in Algorithm~\ref{alg:dt_algo_A}.

\subsection{Application to Private Neural Network Training}\label{subsec:app_nn}
The process of deploying our system for neural network learning is distinct from the process outlined in the previous section for decision tree learning. In central neural network training, after a randomly initialized model of pre-defined structure is created, the following process is used: (1) the dataset $D$ is shuffled and then equally divided into batches, (2) each batch is passed through the model iteratively, (3) a loss function $\mathcal{L}$ is used to compute the error of the model on each batch, (4) errors are then propagated back through the network where an optimizer such as Stochastic Gradient Descent (SGD) is used to update network weights before processing the next batch. Steps (1) through (4) constitute one epoch of learning and are repeated until the model \textit{converges} (stops demonstrating improved performance).

In our system we equate one query to the data parties as one epoch of local learning. That is, each party conducts steps (1) through (4) for one iteration and then sends an updated model to the aggregator. The aggregator then averages the new model weights provided by each party. An updated model is then sent along with a new query for another epoch of learning to each party.

\begin{algorithm}[ht]
\begin{footnotesize}
\caption{Private CNN Learning: Aggregator}\label{alg:nn_algo_A}
\begin{algorithmic}
\STATE \textbf{Input}: Set of data parties $\mathcal{P}$; minimum number of honest, non-colluding parties $t$; noise parameter $\sigma$; learning rate $\eta$; sampling probability $b$; loss function $\mathcal{L}$; clipping value $c$; number of epochs $E$; public key $pk$
\STATE $\bar{t} = n - t + 1$
\STATE Initialize model $M$ with random weights $\theta$;
\FOREACH{$e \in [E]$}
    \STATE Asynchronously query $\mathcal{P}$:
    \STATE \hspace{1em} \textit{train\_epoch}($M$, $\eta$, $b$, $\mathcal{L}$, $c$, $\sigma$, $t$)
    \STATE $\theta_e = $ decrypted aggregate, noisy parameters from $\mathcal{P}$
    \STATE $M \leftarrow \theta_e$
\ENDFOR
\STATE \textbf{return} $M$
\end{algorithmic}
\end{footnotesize}
\end{algorithm}

Each epoch receives the noise parameter $\sigma$ and cost to the overall privacy budget is determined through a separate privacy accountant utility. Just as the decision tree stopping condition was replaced with a pre-set depth the neural network stopping condition of convergence is replaced with a pre-defined number of epochs $E$. 
This process from the aggregator perspective is outlined Algorithm~\ref{alg:nn_algo_A}.

\begin{algorithm}[ht]
\begin{footnotesize}
\caption{Private CNN Learning: Data Party $P_i$}\label{alg:nn_algo_P}
\begin{algorithmic}
\STATE \textbf{procedure} \textsc{train\_epoch}($M$, $\eta$, $b$, $\mathcal{L}$, $c$, $\sigma$, $t$)
\bindent
    \STATE $\theta = $ parameters of $M$
    \FOR {$j \in \{1, 2, ..., \frac{1}{b}\}$}
    \bindent
        \STATE Randomly sample $D_{i,j}$ from $D_i$ w/ probability $b$
        \FOREACH {$d \in D_{i,j}$}
        \bindent
            \STATE $\mathbf{g_j}(d) \leftarrow \triangledown \theta \mathcal{L}(\theta, d)$
            \STATE $\mathbf{\bar{g}_j}(d) \leftarrow \mathbf{g}_j(d) /$ max$\left(1, \frac{||\mathbf{g}_j(d)||_2}{c}\right)$
        \eindent
        \ENDFOR
        \STATE $\mathbf{\bar{g}_j} \leftarrow \frac{1}{|D_{i,j}|}\left(\sum_{\forall d} \mathbf{\bar{g}_j}(d) + \mathcal{N}\left(0, c^2 \cdot \frac{\sigma^2}{t-1}\right)\right)$\;
        \STATE $\theta \leftarrow \theta - \eta \mathbf{\bar{g}_j}$\;
        \STATE $M \leftarrow \theta$
    \eindent
    \ENDFOR
    \STATE \textbf{return} $Enc_{pk}(\theta)$
\eindent
\STATE \textbf{end procedure}
\end{algorithmic}
\end{footnotesize}
\end{algorithm}

At each data party we deploy code to support the process detailed in Algorithm~\ref{alg:nn_algo_P}. To conduct a complete epoch of learning we follow the approach proposed in~\cite{abadi2016deep} for private centralized neural network learning. This requires a number of changes to the traditional learning approach. Rather than shuffling the dataset into equal sized batches, a batch is randomly sampled for processing with sampling probability $b$. An epoch then becomes defined as the number of batch iterations required to process $|D_i|$ instances. Additionally, parameter updates determined through the loss function $\mathcal{L}$ are clipped to define the sensitivity of the neural network learning to individual training instances. Noise is then added to the weight updates. Once an entire epoch is completed the updated weights can be sent back to the aggregator.

\subsection{Application to Private Support Vector Machine Training}\label{subsec:app_svm}

Finally, we focus on the classic $\ell_2$-regularized binary linear SVM problem with hinge loss, which is given in the following form:
\begin{equation}\label{svm-loss}
\mathcal{L}(w):=\frac{1}{|D|}\sum_{\forall (x_i,y_i)\in D}\max\{0,1-y_i\langle w, x_i\rangle\} + \lambda\|w\|_2^2,
\end{equation}
where $(x_i,y_i)\in \mathbb{R}^d \times \{-1,1\}$ is a feature vector, class label pair, $w\in \mathbb{R}^d$ is the model weight vector, and $\lambda$ is the regularized coefficient.

From the aggregator perspective, specified in Algorithm~\ref{alg:svm-agg}, the process of SVM training is similar to that of neural network training. Each query to the data parties is defined as $K$ epochs of training. Once query responses are received, model parameters are averaged to generate a new support vector machine model. This new model is then sent to the data parties for another $K$ epochs of training. We again specify a pre-determined number of epochs $E$ to control the number of training iterations.
\begin{algorithm}[ht]
\begin{footnotesize}
\caption{Private SVM Learning: Aggregator}\label{alg:svm-agg}
\begin{algorithmic}

\STATE \textbf{Input}: Set of data parties $\mathcal{P}$; minimum number of honest, non-colluding parties $t$; noise parameter $\sigma$; learning rate $\eta$; loss function $\mathcal{L}$; clipping value $c$; number of epochs $E$; number of epochs per query $K$, public key $pk$
\STATE $\bar{t} = n - t + 1$
\STATE Initialize model $M$ with random weights $w$;
\FOREACH{$e \in [E/K]$}
    \STATE Asynchronously query $\mathcal{P}$:
    \STATE \hspace{1em} \textit{train\_epoch}($M$, $\eta$, $K$, $\mathcal{L}$, $c$, $\sigma$, $t$)
    \STATE $\theta_e = $ decrypted aggregate, noisy parameters from $\mathcal{P}$
    \STATE $M \leftarrow \theta_e$
\ENDFOR
\STATE \textbf{return} $M$
\end{algorithmic}
\end{footnotesize}
\end{algorithm}

To complete an epoch of learning at each data party, we iterate through each instance in the local training dataset $D_i$. We again deploy a clipping approach to constrain the sensitivity of the updates. The model parameters are then updated according to the loss function $\mathcal{L}$ as well as the noise parameter. The process conducted at each data party for $K$ epochs of training in response to an aggregator query is outlined in Algorithm~\ref{alg:svm-party}.

\begin{algorithm}[ht]
\begin{footnotesize}
\caption{Private SVM Learning: Data Party $P_i$}\label{alg:svm-party}
\begin{algorithmic}
\STATE \textbf{procedure} \textsc{train\_epoch}($M$, $\eta$, $K$, $\mathcal{L}$, $c$, $\sigma$, $t$)
\bindent
    \STATE $w = $ parameters of $M$
        \FOREACH {$(x_i,y_i) \in D$}
        \bindent
            \STATE ${x}_i \leftarrow x_i /$ max$\left(1, \frac{||x_i||_2}{c}\right)$ 
        \eindent
        \ENDFOR
        \FOR {$k \in \{1, 2, ..., K\}$}
        \bindent
        \STATE $\mathbf{g}(D) \leftarrow \triangledown w \mathcal{L}(w, D)$
        \STATE $\mathbf{\bar g} \leftarrow  \mathbf{g}(D) + \mathcal{N}\left(0, \frac{\sigma^2}{t-1}\right)$\;
        \STATE $w \leftarrow w - \eta \mathbf{\bar{g}}$\;
        \STATE $M \leftarrow w$
    \eindent
        \ENDFOR
    \STATE \textbf{return} $Enc_{pk}(w)$
\eindent
\STATE \textbf{end procedure}
\end{algorithmic}
\end{footnotesize}
\end{algorithm}
\subsection{Expanding the Algorithm Repository}

Beyond the three models evaluated here, our approach can be used to extend any differentially private machine learning algorithm into a federated learning environment. We demonstrate the flexibility of our system through 3 example algorithms which are of broad interest and significantly different. The task of generating and deploying our system for each algorithm, however, is non-trivial. First, a DP version of the algorithm must be developed. Second, this must be written as a series of queries. Finally, each query must have an appropriate aggregation procedure. Our approach may then be applied for accurate, federated, private results.

Due to our choices to use the threshold Paillier cryptosystem in conjunction with an aggregator, rather than a complex SMC protocol run by the parties themselves, we can provide a streamlined interface between the aggregator and the parties. Parties need only answer data queries with encrypted, noisy responses and decryption queries with partial decryption values. Management of the global model and communication with all other parties falls to the aggregator, therefore decreasing the barrier to entry for parties to engage in our federated learning system. Figure~\ref{fig:train_time} demonstrates the impact of this choice as our approach is able to effectively handle the introduction of more parties into the federated learning system without the introduction of increased encryption overhead.

Another issue in the deployment of new machine learning training algorithms is the choice of algorithmic parameters. Key decisions must be made when using our system and many are domain-specific. We aim to inform such decisions with our analysis of trade-offs between privacy, trust and accuracy in Section~\ref{subsubsec:variation_settings}, but note that the impact will vary depending on the data and the training algorithm chosen. While our system will reduce the amount of noise required to train any federated ML algorithm, questions surrounding what impact various data-specific features will have on the privacy budget are algorithm-specific. For example, Algorithm~\ref{alg:dt_algo_A} demonstrates how, in decision tree training, the number of features and classes impact the privacy budget at each level. Similarly, Algorithms~\ref{alg:nn_algo_P} and~\ref{alg:svm-party} show the role of norm clipping in neural network and SVM learning. In neural networks, this value not only impacts noise but will also have a different impact on learning depending on the size of the network and number of features.

\section{Related Work}
Our work relates to both the areas of FL as well as privacy-preserving ML.
Existing work can be classified into three categories:
trusted aggregator, local DP, and cryptographic.

\paragraph{Trusted Aggregator}
Approaches in this area trust the aggregator to obtain data in plaintext or add noise.
~\cite{abadi2016deep} and~\cite{jagannathan2009practical} propose differentially private ML systems,
but do not consider a distributed data scenario, thus requiring a central party.
In~~\cite{zhang2011distributed}, the authors develop a distributed data mining system with DP but show significant accuracy loss and require 
a trusted aggregator to add noise.

Recently, \cite{papernot:pate:2018} presented PATE, an ensemble approach to private learning wherein several ``teacher'' models are independently trained over local datasets. A trusted aggregator then provides a DP query interface to a ``student'' model that has unlabelled public data (but no direct access to private data) and obtains labels through queries to the teachers. While we have proposed a federated learning (FL) approach wherein one global model is learned over the aggregate of the parties' datasets, the PATE method develops an ensemble model with independently trained base models using local datasets. Unlike the methods we evaluate, PATE assumes a fully trusted party to aggregate the teachers' labels; focuses on scenarios wherein each party has enough data to train an accurate model, which might not hold, e.g., for cellphone users training a neural network; and assumes access to publicly available data, an assumption not made in our FL system. Models produced from our FL system learn from all available data, leading to more accurate models than the local models trained by each participant in PATE (Figure 4b in~\cite{papernot:pate:2018} demonstrates the need for a lot of parties to achieve reasonable accuracy in such a setting).

\paragraph{Local Differential Privacy}
\cite{shokri2015privacy} presents a distributed learning system using DP without a central trusted party. However, the DP guarantee is per-parameter and becomes meaningless for models with more than a small number of parameters.

\paragraph{Cryptographic Approaches}
\cite{shi2011privacy} presents a protocol to privately aggregate sums over multiple time periods. Their protocol is designed to allow participants to periodically upload encrypted values to an oblivious aggregator with minimum communication costs. Their approach however has participants sending in a stream of statistics and does not address FL or propose an FL system. 
Additionally, their approach calls for each participant to add noise independently. As our experimental results show, allowing each participant to add noise in this fashion results in models with low accuracy, making this approach is unsuitable for FL. In contrast, our approach reduces the amount of noise injected by each participant by taking advantage of the additive properties of DP and the use of threshold-based homomorphic encryption to produce accurate models that protect individual parties' privacy.

In~\citep[\S B]{bonawitz2017practical} the authors propose the use of multiparty computation to securely aggregate data for FL. 
The focus of the paper is to present suitable cryptographic techniques to ensure that the aggregation process can take place in mobile environments. 
While the authors propose FL as motivation, no complete system is developed with ``a detailed study of the integration of differential privacy, secure aggregation, and deep learning'' remaining beyond the scope.

\cite{beimelDistributed2008} provides a theoretical analysis on how differentially private computations could be done in a federated setting for single instance operations using either secure function evaluation or the local model with a semi-trusted curator.
By comparison, we consider multiple operations to conduct FL and provide empirical evaluation of the FL system.
\cite{narayan2012djoin} proposes a system to perform differentially private database joins. 
This approach combines private set intersection with random padding, but cannot be generally applied to FL. 
In~\cite{pettai2015combining} the authors' 
protocols are tailored to inner join tables and counting the number of values in an array. In contrast, we propose an accurate, private FL system for predictive model training. 

Dwork et al.~\cite{dwork2006our} present a distributed noise generation scheme and focus on methods for generating noise from different distributions. This scheme is based on secret sharing, an MPC mechanism that requires extensive exchange of messages and entails a communication overhead not viable in many federated learning settings.

\cite{chase2017private} proposes a method to train neural networks in a private collaborative fashion by combining MPC, DP and secret sharing assuming non-colluding honest parties. In contrast, our system prevents privacy leakages even if parties actively collude.

Approaches for the private collection of streaming data, including ~\cite{7286780,ChanSS12,acs2011have,rastogi2010differentially}, aim to recover computation when one or more parties go down. Our system, however, enables private federated learning which allows for checkpoints in each epoch of training. The use of threshold cryptography also enables our system to decrypt values when only a subset of the participants is available.

\section{Conclusion}

In this paper, we present a novel approach to perform FL that combines DP and SMC to improve model accuracy while preserving provable privacy guarantees and protecting against  
extraction attacks and collusion threats.
Our approach can be applied to train different ML models in a federated learning fashion for varying trust scenarios.
Through adherence to the DP framework we are able to guarantee overall privacy from inference of any model output from our system as well as any intermediate result made available to $\mathcal{A}$ or $\mathcal{P}$.
SMC additionally guarantees any messages exchanged without DP protection are not revealed and therefore do not leak any private information.
This provides end-to-end privacy guarantees with respect to the participants as well as any attackers of the model itself.
Given these guarantees, models produced by our system can be safely deployed to production without infringing on privacy guarantees.

We demonstrated how to apply our approach to train a variety of ML models and showed that it out-performs existing state-of-the art techniques for FL.
Our system provides significant gains in accuracy when compared to a na\"ive application of state-of-the-art differentially private protocols to FL systems.

For a tailored threat model, we propose an end-to-end private federated learning system which uses SMC in combination with DP to produce models with high accuracy. 
As far as we know this is the first paper to demonstrate that the application of these combined techniques allow us to maintain this high accuracy at a given level of privacy over different learning approaches. In the light of the ongoing social discussion on privacy, this proposed approach provides a novel method for organizations to use ML in applications requiring high model performance while addressing privacy needs and regulatory compliance.

\bibliography{aisecbib}
\bibliographystyle{ACM-Reference-Format}






\end{document}